\documentclass[conference,letterpaper]{IEEEtran}
\makeatletter
\renewcommand*{\thanks}[1]{%
  \footnotemark
  \protected@xdef\@thanks{\@thanks
    \protect\footnotetext[\arabic{footnote}]{#1}}%
}
\makeatother

\ifCLASSINFOpdf
\else
\fi
\usepackage{todonotes}
\usepackage{caption}
\usepackage{subcaption }
\usepackage{amssymb}
\usepackage{amsmath}

\usepackage{blindtext}

\begin{document}
\newcommand{\argmax}{\operatornamewithlimits{argmax}}
\newcommand{\argmin}{\operatornamewithlimits{argmin}}

% Tensorflow color map: 
% http://www.mulinblog.com/a-color-palette-optimized-for-data-visualization/
%\definecolor{tf_red}{HTML}{F15854}
%\definecolor{tf_blue}{HTML}{5DA5DA}
%\definecolor{tf_orange}{HTML}{FAA43A}
%\definecolor{tf_green}{HTML}{60BD68}
%\definecolor{tf_pink}{HTML}{F17CB0}
%\definecolor{tf_brown}{HTML}{B2912F}
%\definecolor{tf_purple}{HTML}{B276B2}
%\definecolor{tf_yellow}{HTML}{DECF3F}
%\definecolor{tf_gray}{HTML}{4D4D4D}

\title{Imitation Learning for Vision-based\\ Lane Keeping Assistance}
\author{Christopher Innocenti$^*$, Henrik Lind\'{e}n$^*$, Ghazaleh Panahandeh, Lennart Svensson, Nasser Mohammadiha
\thanks{
This paper will appear in the Proceedings of  the International Conference on Intelligent Transportation Systems (ITSC), 2017. Citation: C. Innocenti, H. Lind\'{e}n, G. Panahandeh, L. Svensson, and N. Mohammadiha, "Imitation Learning for Vision-based Lane Keeping Assistance​", in Proc. of the International Conference on Intelligent Transportation Systems (ITSC), 2017.  \newline
Christopher Innocenti and Henrik Lind\'{e}n worked on this project while doing their Master thesis at Volvo Cars. Ghazaleh Panahandeh and Nasser Mohammadiha are with Zenuity AB, Gothenburg, Sweden. Lennart Svensson is with Chalmers University of Technology,
Gothenburg, Sweden.
Email$^*$: $\{$chrinnoc, henke.linden$\}$@gmail.com}
% \author{\IEEEauthorblockN{%
% Christopher Innocenti\IEEEauthorrefmark{1},
% Henrik Lind\'{e}n\IEEEauthorrefmark{1},
% Ghazaleh Panahandeh\IEEEauthorrefmark{1}\IEEEauthorrefmark{2},
% Lennart Svensson\IEEEauthorrefmark{3},
% Nasser Mohammadiha\IEEEauthorrefmark{1}\IEEEauthorrefmark{3}
% %Author3 \IEEEauthorrefmark{2} and
% %Author4 \IEEEauthorrefmark{3}
% }
% \IEEEauthorblockA{\IEEEauthorrefmark{1}Zenuity AB, 
% \IEEEauthorrefmark{2}Volvo Car Corporation, \IEEEauthorrefmark{3}Chalmers University of Technology\\
% Gothenburg, Sweden\\
% Email: $\{$chrinnoc, henke.linden$\}$@gmail.com\\
% }

%\IEEEauthorblockA{\IEEEauthorrefmark{2}Starfleet Academy, San Francisco, California 94103\\
%Email: $\{$kirk,scott$\}$@starfleet.com}

%\IEEEauthorblockA{\IEEEauthorrefmark{3}Tyrell Inc., 123 Replicant Street, Los Angeles, California 90210\\
%Email: tyrell@electricsheep.com}
}

\maketitle

% =======================================================
% =====> ABSTRACT
% =======================================================

\begin{abstract}
This paper aims to investigate direct imitation learning from human drivers for the task of lane keeping assistance in highway and country roads using grayscale images from a single front view camera. The employed method utilizes convolutional neural networks (CNN) to act as a policy that is driving a vehicle. The policy is successfully learned via imitation learning using real-world data collected from human drivers and is evaluated in closed-loop simulated environments, demonstrating good driving behaviour and a robustness for domain changes. Evaluation is based on two proposed performance metrics measuring how well the vehicle is positioned in a lane and the smoothness of the driven trajectory.
% and the possibility for rapid and inexpensive prototyping. %From the results obtained, it is then argued that such models can serve as powerful tools for autonomous driving with comparatively simple sensor equipment. 

\end{abstract}

% =======================================================
% =====> INTRODUCTION
% =======================================================

\section{Introduction}

The automotive industry is currently a very active area for machine learning, and most major companies are dedicating huge effort into creating systems capable of Autonomous Driving (AD). In many AD applications, it is common that raw data from vehicle sensors, such as that from a camera, radar, or lidar is processed and fused in order to provide an accurate description of the environment in which the vehicle is driven. This description of the environmental state can then be utilized for path planning, adaptive speed control, or collision avoidance systems. That is, commonly, autonomous driving agents consist of some large system of control logic that makes use of acquired data in some engineered way in order to safely control the vehicle. 

While the manual engineering of such control logic enables great flexibility, in the sense that the acquired data can be used in a more controlled way, %almost any human made system structure,
it may be unlikely that it results in an optimal system. In addition, creating, and making good use of features from the raw data requires top knowledge within the respective fields which may be difficult to acquire, which in turn might drive up the cost of manual engineering considerably. A way to remedy the issue of manual engineering is to employ a deep learning (DL) approach as it is data-driven and thus has the capability to learn relevant features from data without human supervision. %Because deep learning models can operate on raw input data, they completely remove the feature engineering requirement and might have the potential to work in an end--to--end fashion.

The main contributions of this work include: 
(a) successful policy learning for steering a vehicle using only grayscale images from a single front facing camera without any data augmentation by means of direct imitation learning, 
(b) application of domain adaption to show that a model learning from real data performs well in simulated environments, allowing for rapid and inexpensive model evaluation in favor of initial hardware-in-the-loop (HIL) and/or real vehicle tests, and
(c) demonstrating that a diverse enough dataset consisting solely of front facing images seems sufficient in order to teach a vehicle control model to remain well positioned in a lane, with the ability to recover from disturbances.

\subsection{Motivation and Related Work}\label{sec:relwork}
In recent years, data-driven methods such as DL has sharply increased in popularity with many impressing results within the AD field to date. As early as 1989, the project ALVINN \cite{alvinn-1989} investigated whether a simple neural network could be sufficient for steering a vehicle. Compared to the size of some of the networks employed for classification etc. today, the ALVINN system was tiny as it only used a very small fully connected feed forward network to directly map camera and lidar input from the road ahead into appropriate steering actions. Although the system was trained mostly on artificial data, it managed to drive a 400-meter path through a wooded area of the Carnegie Mellon University campus under sunny conditions with a speed of 0.5m/s, indicating that neural networks could be a potential solution to AD.

Further, in 2005, an effort for autonomous driving of a small RC-vehicle in an off-road environment, in the form of obstacle avoidance, was performed \cite{dave-2005}. The project, called DAVE, utilized a more advanced neural network architecture including convolutional layers to learn how to avoid obstacles when driving in the terrain. The training was performed by using real world data captured from the vehicle front mounted stereo camera, and the steering commands that the human expert had used when collecting the data. Based on an evaluation set, the DAVE system performed quite poorly with almost 37\% error, however, the authors reported much better performance in practice, claiming that the use of an evaluation set was not representative.

Recently, in 2016, researchers at Nvidia developed a deep learning based system \cite{dave2-2016} for lane following to be trained end--to--end as well. Their project, named DAVE2, used an even more advanced CNN than the one used in the original DAVE project and was trained on a larger dataset. The data consisted of everyday driving on public roads, captured from three front mounted cameras, although at inference time the system operated on images from a single camera. As reported by Nvidia, their vehicle managed to drive autonomously 98$\%$ of the time in their trials on highways.

%\subsection{Problem Formulation}
Possible limitations with the described systems are that they all rely on a multitude of sensors, e.g. \cite{alvinn-1989} used both camera and lidar, \cite{dave-2005} used stereo cameras and \cite{dave2-2016} used three cameras in order to learn robust driving behaviour (although \cite{dave2-2016} uses only one camera once trained). However, because cameras provide very rich information and are relatively cheap sensors, it is interesting to investigate whether robust driving behaviour can be learned by means of imitation learning using only a single front view camera. %Further, using just a single front view camera has been suggested not to work in \cite{dave2-2016} without data augmentation or otherwise including scenarios where the vehicle recovers from bad situations into the dataset used for training. 

%\todo{here}
%{\color{red}you listed some of the works done previously, however it is not clear what is your contributions compared to what you listed in this subsection. You may rewrite your main contribution here in connection with the related works here. Starting by e.g. "Our contribution in this paper compared to ...".}

% =======================================================
% =====> METHODS
% =======================================================

\section{Imitation Learning for Vehicle Control}\label{sec:imitlearn}

\begin{figure}
    \centering
    \includegraphics[width=\columnwidth]{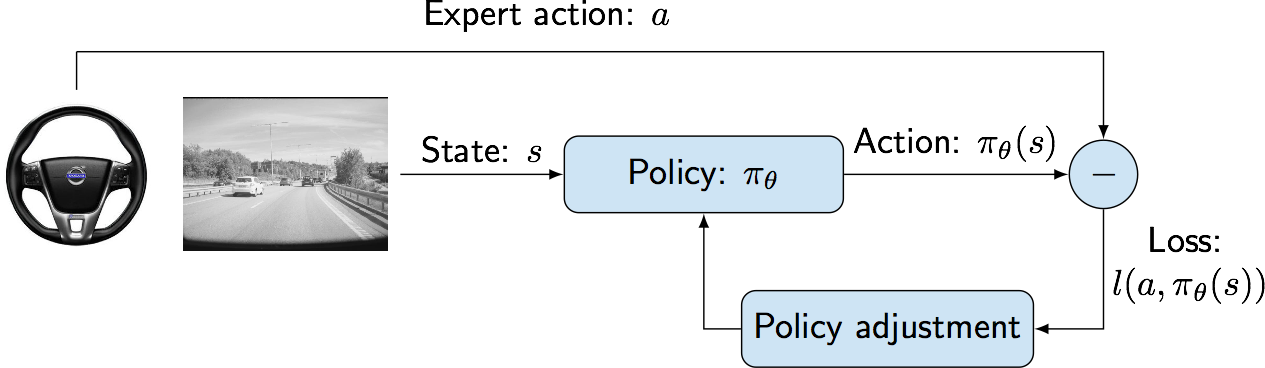}
    %\resizebox{\columnwidth}{!}{\input{figures/imlearn.tex}}
    \caption{Illustration of direct imitation learning from expert state--action pairs $(s,a)$ where the vehicle steering policy $\pi_{\theta}$ is updated through backpropagation based on the loss $l(a,\pi_{\theta}(s))$ between the action taken under policy and that of the expert.}
    \label{fig:imlearn}
\end{figure}

Imitation learning (IL) is a method for learning a mapping $\pi$, denoted a \textit{policy}, from states $s$ to actions $a$. IL requires (and is based on the assumption) that there exists an expert whose behaviour can be imitated in order to learn a desired behaviour. In this paper, the task of learning a policy for inferring steering actions from front view camera images was done by means of direct imitating learning (DIL) based on the behaviour of human drivers. 
%This corresponded to learning a distribution $\pi_{\theta}(a|s)$ of steering actions $a$ conditioned on a state observation $s$ (an image from the vehicle front view camera). 
For this work, $\pi_{\theta}$ was modelled as a deterministic policy using a CNN with parameters $\theta$ to approximate driving behaviour.
%\begin{equation}
%   a^* = \argmax_a \pi_{\theta}(a|s).
%\end{equation}

For the remainder of this section, the concept of DIL is further explained in Section~\ref{sec:prob_form}, whereas an overview of the dataset used for training is given in Section Section~\ref{sec:dataset}. Finally, Section~\ref{sec:arch} and \ref{sec:train} provide descriptions of our CNN architecture and the training procedure.

\subsection{Direct Imitation Learning}\label{sec:prob_form}

The task of learning a behaviour directly from expert state--action pairs $(s,a)$ is commonly known as DIL or \textit{behavioural cloning} \cite{imitation2017, togelius2007}. Application of DIL is thus analogous to that of supervised learning with the main difference that input--output pairs, generally represented by feature--label pairs $(x,y)$, are examples of states and actions. In general, once a policy is trained using DIL, policy refinement is often said to be required in order to learn a more robust behaviour. This involves a post learning stage where the policy is refined by e.g. reinforcement learning (RL) techniques or data aggregation algorithms \cite{dave2-2016,imitation2017,dagger2011}. %For DIL however, no post learning stage is performed. 

By not including any policy refinement, $\pi_{\theta}$ is directly (and only) learned from the distribution $d_{\pi^*}(s,a)$ of states, $s$, and actions, $a$, encountered under an expert policy $\pi^*$, i.e. the states and actions observed from human drivers. The objective is then to learn a policy parametrization $\theta^*$ which minimizes the expected loss $l(a,\pi_{\theta}(s))$ under $(s,a) \sim d_{\pi^*}$ as
\begin{equation}
 \theta^* = \argmin_{\theta} \mathbb{E}_{(s,a) \sim d_{\pi^*}}\left[l(a,\pi_{\theta}(s))\right].
\end{equation}
For this work, the loss corresponds to the squared error between actions and the actions inferred by the policy at the corresponding state as
\begin{equation}
 l(a,\pi_{\theta}(s)) = (a - \pi_{\theta}(s))^2.
\end{equation}
The expected loss is approximated as the mean squared error as
%the loss corresponded to the mean squared error (MSE) 
%for batches 
%of the state--action pairs of the dataset and the corresponding actions inferred under the policy as
\begin{equation}
    \mathbb{E}_{(s,a) \sim d_{\pi^*}}\left[l(a,\pi_{\theta}(s))\right] \approx \frac{1}{N}\sum_{i=1}^N (a_i - \pi_{\theta}(s_i))^2 \label{eq:mse}
\end{equation}
where $N$ is the number of state--action pairs in the dataset used for the learning procedure.

Figure~\ref{fig:imlearn} illustrates the policy learning process where the state representation $s$ corresponds to an image of a vehicle's front-view-camera and the expert action $a$ corresponds to the steering action applied by a human driver at the corresponding point in time. The policy learning was performed as iterative parameter updates by backpropagation of the MSE gradient with respect to $\theta$ using stochastic gradient descent.

%In this work, a CNN model is trained by direct imitation learning to learn expert behaviour (actions of human drivers) and represents a policy $\pi_{\theta}$ with trainable parameters $\theta$. 
%This section describes the dataset used, the model architecture, and the learning procedure.

% -------------------------------------------------------
% -----> DATASET
% -------------------------------------------------------

\subsection{Dataset} \label{sec:dataset}

\begin{figure}[t]
\centering
\includegraphics[width=\columnwidth]{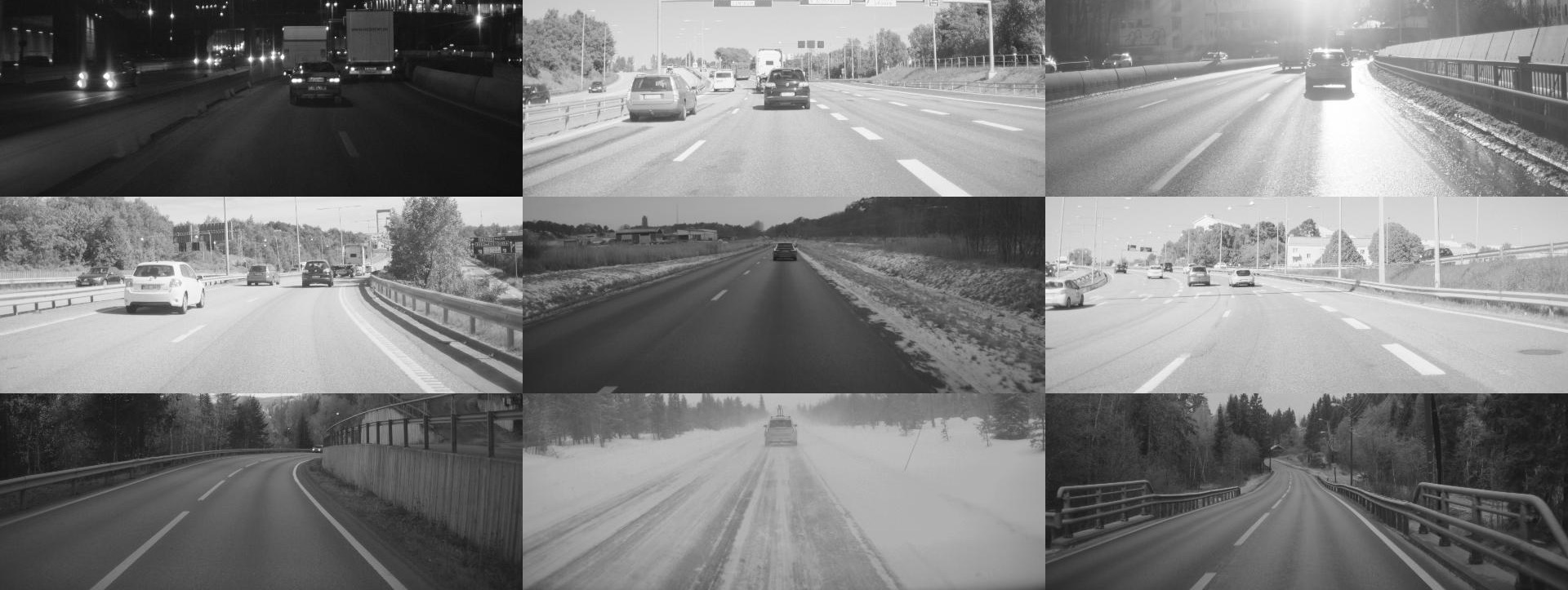}
\caption{A selection of (cropped) images used for training captured from the front view camera of a test vehicle illustrating a variety of lighting and weather conditions.}
\label{fig:cads4examples}
\end{figure}

%For the performed study, the data to be used for training the model was supplied from Volvo Cars. 
The data used for training was supplied from Volvo Car Corporation (VCC). The data that was made available consisted of a large number of logfiles. Each logfile was recorded a few minutes at a time during various expeditions VCC had undertaken over the last five years. An expedition usually lasted a couple of days, driving the vehicle from city A to another city B, in different parts of Europe. Naturally, the variation in the time of capture and the actual driving resulted in the data having a high variability in terms of road and weather conditions etc.

%The available data consisted of a large number of log files collected over the last 5 {\color{red} data from xc90 over 5 years!?} years of expeditions under different road and weather conditions.
Each of the numerous logfiles contained for example, the video-feed from the test vehicles' front mounted cameras, but also the state of many of the vehicles' sensors.
For the purposes of this project, image frames were sampled from the videos at 20 Hz coupled with their corresponding steering wheel angles (SWA) as acted out by human drivers for each frame. Figure~\ref{fig:cads4examples} shows a selection of images captured from various expeditions under a variety of road and weather conditions. In total, roughly 2.5 million image--steering (state--action) pairs were extracted from the logfiles.

\begin{figure}[t]
\centering
\includegraphics[width=0.4\columnwidth]{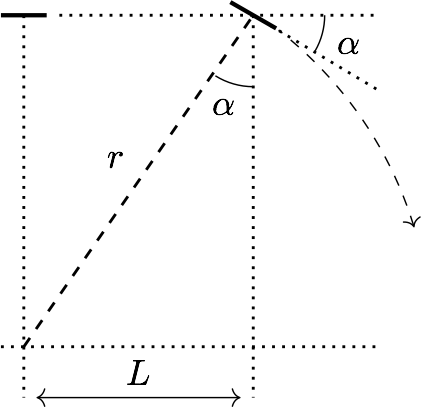}
%\resizebox{0.4\columnwidth}{!}{\input{figures/bicycle.tex}}
\caption{A bicycle representation for calculation of the turning radius $r$ of a vehicle with wheel base $L$ and front wheel angle $\alpha$.}
\label{fig:bicycle}
\end{figure}

It was desired that the policy should be general and capable of operating on various vehicle platforms. However, as nearly every car model, or at least every brand has different dimensions (wheelbase etc.), range and conversion rates from its steering wheel to the front wheel angles, using the raw recording of the SWA as an action could be problematic. Doing so would mean that a trained policy might have become dependent on the actual vehicle used to collect the log-data and not work properly when controlling other vehicle models. To circumvent such an issue, a transformation of the SWA was made. Instead of using the SWA directly, it was transformed to the vehicle's turn radius $r$ and subsequently to a measure of curvature $\frac{1}{r}$. The transformation was done by assuming a simple bicycle model for the vehicle as seen in Figure~\ref{fig:bicycle}.

% -------------------------------------------------------
% -----> METHODOLOGY
% -------------------------------------------------------
%\subsection{Methodology}

\subsection{Model Architecture}\label{sec:arch}

\begin{figure}[t]
\centering
\includegraphics[width=\columnwidth]{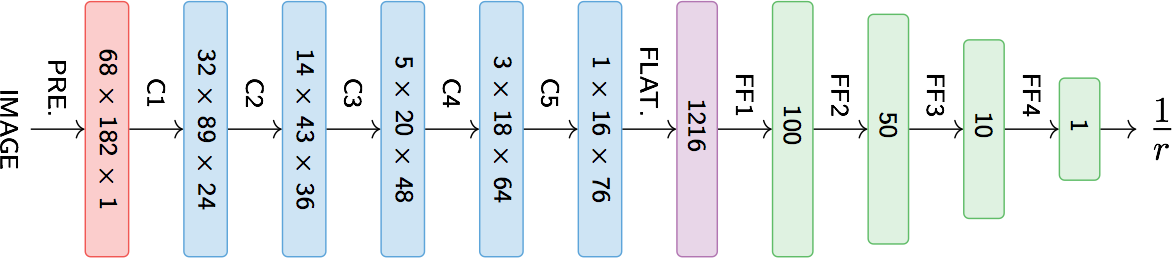}
%\resizebox {\columnwidth} {!} {\input{figures/netarch.tex}}
\caption{Model architecture consisting of a static preprocessing layer, five convolutional layers, one reshaping layer, and four fully connected layers. The preprocess is denoted by ``PRE", the convolutions by ``C1,$\ldots$,C5", the reshaping by ``FLAT", and the fully connected layers by ``FF1,$\ldots$,FF4".}
\label{fig:net-architecture}
\end{figure}

The proposed CNN model consists of approximately $265\times10^3$ parameters with an architectural layout similar to Nvidia's PilotNet \cite{dave2-2016}. Figure~\ref{fig:net-architecture} illustrates the model architecture, the applied operations, and the feature map sizes at each layer. 

Image inputs are initially preprocessed by: (a) cropping away 35\% of the top and 15\% of the bottom of the images, (b) down sampling a factor 3.5 using the bi-linear transform and (c) normalizing by per image subtracting the mean and dividing by standard deviation. The motivation for cropping the images is that the original images contained some redundant information such as the hood of the vehicle and an excessive amount of sky. The images also had some vignetting effects that didn't provide any useful information. Both the cropping and the downsampling also serve the purpose of reducing the number of pixels in the input and in turn reducing the computational complexity of the network.

Five layers of strided 2D convolution with ELU activations are then applied to the processed images. The first three conv layers consists of 24, 36, and 48, $5\times 5$ kernels applied with $2\times 2$ strides respectively. The following two conv layers consists of 64 and 76, $3\times 3$ kernels applied with $1\times 1$ strides. The feature maps obtained from the final convolutional layer is then reshaped to vector representation of size 1216 and followed by three fully connected layers with ELU activations. These consist of 100, 50, and 10 hidden units respectively. Lastly, the output layer consists of a single unit with identity activation representing the curvature.

\subsection{Model Training}\label{sec:train}

Training is performed in a supervised setting on the aforementioned dataset using the mean squared error loss between the steering action inferred by the model and that of the expert (the human driver) as given by Equation~\ref{eq:mse}. No data augmentations (shifts, distortions etc.) are applied during training, however, it was quickly noted that when driving a vehicle on country or highway roads, the actual turning of the wheels is seemingly quite limited. Because the data used in this work came from such driving scenarios, it naturally showed the same characteristics. That is to say, the case of driving straight was overrepresented.

\begin{figure}
    \centering
    \includegraphics[width=0.5\columnwidth]{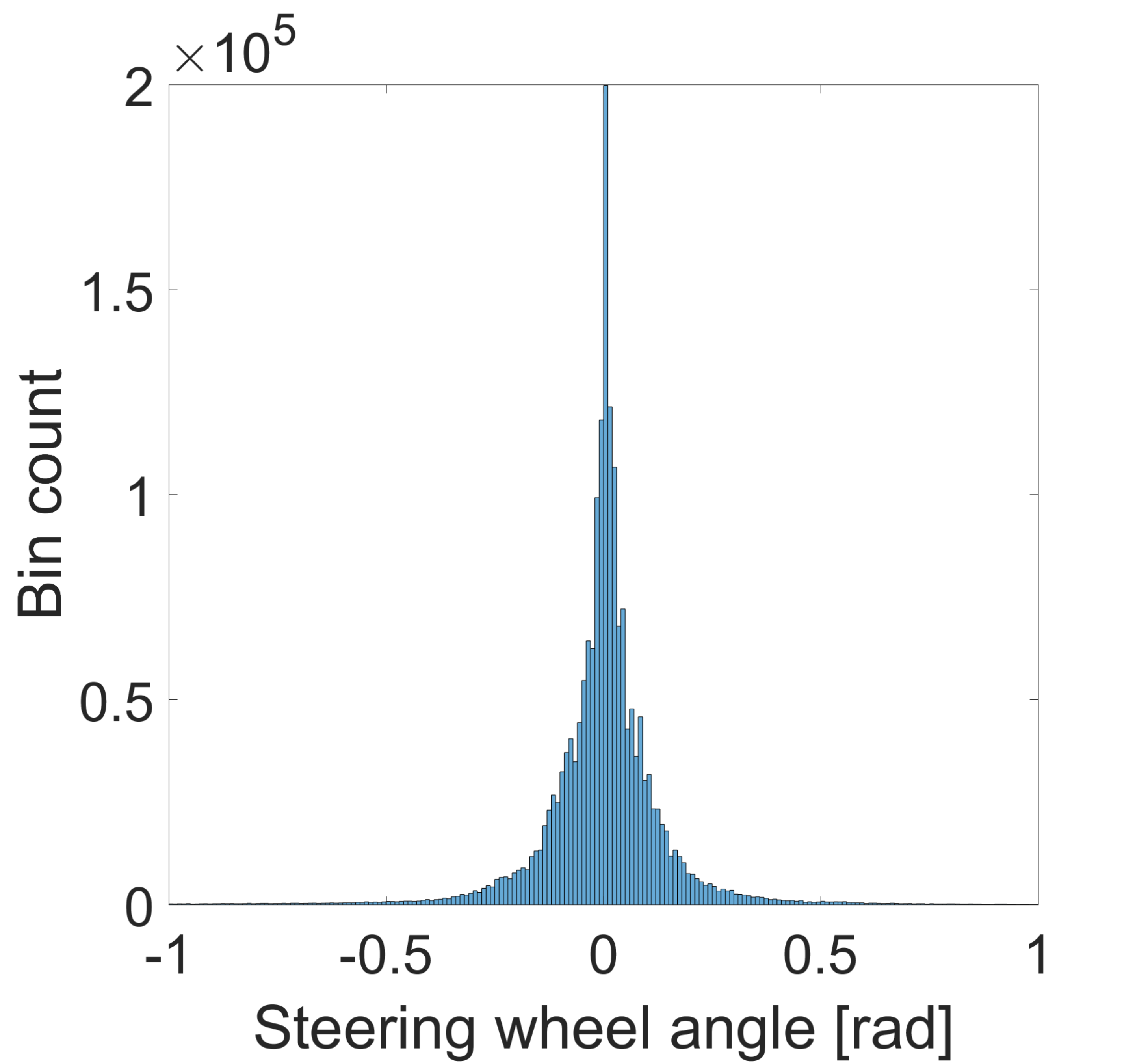}~
    \includegraphics[width=0.5\columnwidth]{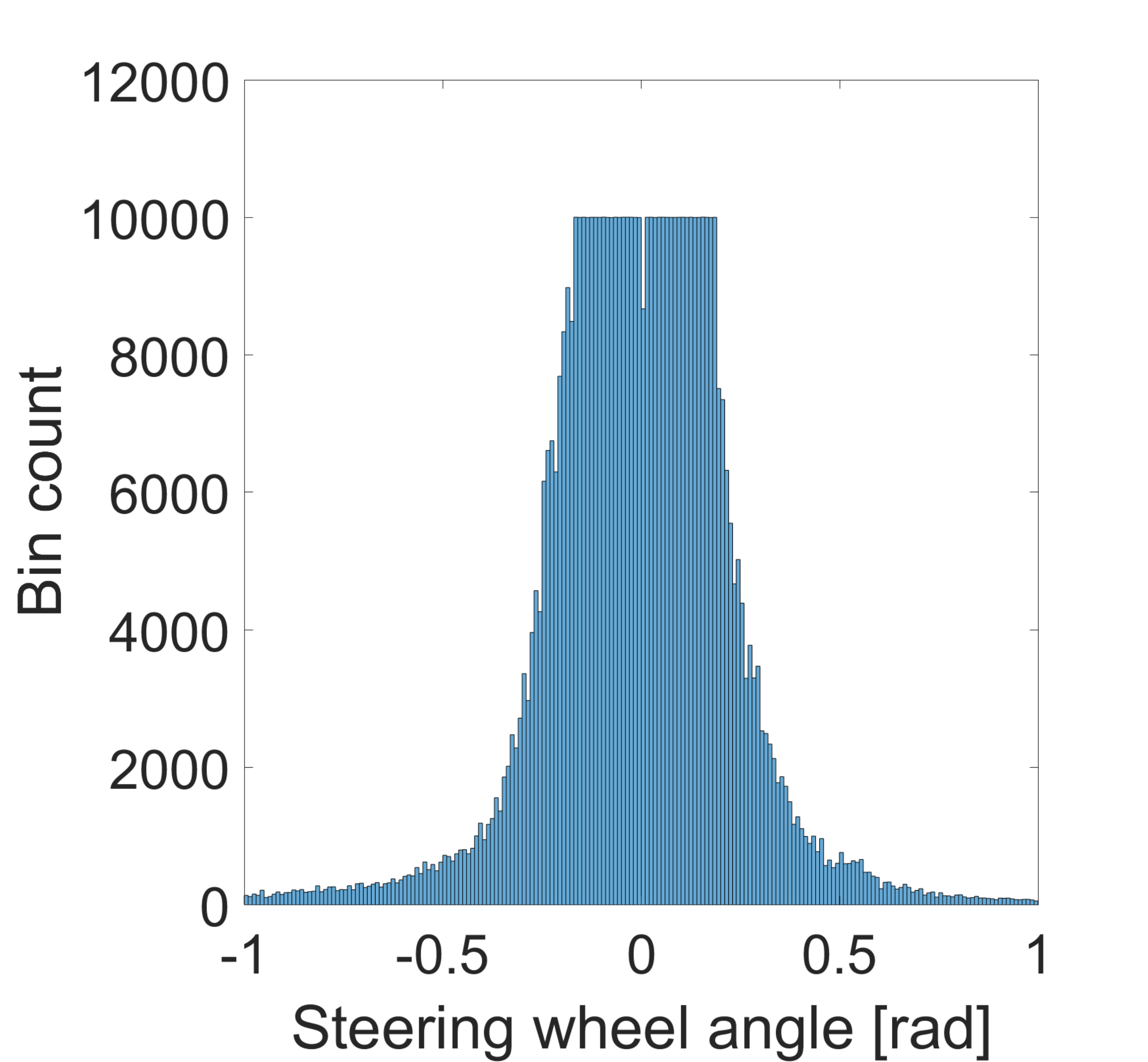}
    \caption{Histograms (in the significant range $\pm1$ rad) of steering wheel angles of the initial dataset (left) and pruned dataset (right).}
    \label{fig:hists}
\end{figure}

To ameliorate the issue of straight driving being overrepresented, the dataset was pruned to get a more uniform spread over the whole range of possible steering actions. The pruning procedure was performed by forming a histogram of the SWAs with 18000 bins in the range $\pm9$ rad (maximum SWA) where the maximum bin-count was set to 10000. Essentially, this meant that some image-steering pairs were discarded from the initial dataset. To maintain diversity, larger expeditions that contained more data points were pruned more than smaller ones. After the pruning process, the final dataset amounted to roughly 1.4 million image--steering pairs, or approximately 27h of driving. Figure~\ref{fig:hists} depicts histograms of the significant range, $\pm1$ rad, of the SWA data before and after the pruning procedure.

With the data properly adjusted, the optimization of the network was done using stochastic gradient descent with the Adam optimization scheme~\cite{adam-2015} having an initial learning rate of 0.0001. The batch size was set to $n = 64$ and dropout~\cite{dropout-2014} was applied to the first three fully connected layers with keep probability $p = 0.5$. The training process was executed for approximately $2\times10^{5}$ batches, corresponding to roughly 9 epochs of the dataset.

% -------------------------------------------------------
% -----> EVALUATION SETUP
% -------------------------------------------------------
\section{Policy Evaluation Setup}

Evaluation of the resulting policy, trained from Volvo expedition data, 
%as described in Section~\ref{sec:imitlearn}, 
was done by means of closed loop simulation in virtual environments. The choice of evaluating the proposed method in simulations, rather than using a real test vehicle, was made since it was deemed a more cost efficient and simple approach. The environments used was IPG CarMaker~\cite{IPG} and the Unity game engine~\cite{Unity5-6}. For the simulations, a virtual front view camera streamed images in real time, which the learned policy mapped to appropriate steering actions acted out by the simulated driver/vehicle in closed loop. 

Quantitative evaluation of the policy was based on two performance metrics, lane positioning and level of discomfort, which were estimated
%and was done by letting the policy act as a simulated driver 
for a variety of test scenarios in the IPG CarMaker environment. The Unity game engine was primarily used for empirical evaluation and to diversify the different scenarios and domains used for evaluation. 

The remainder of this section provides the definitions of the proposed performance metrics in Section~\ref{sec:metrics} and descriptions of the different test scenarios in Section~\ref{sec:tests}.

\subsection{Performance Metrics}\label{sec:metrics}
Rating a system whose performance largely is subjective, as is indeed the case with driving, is not trivial by any means. There are likely many different factors that characterize ``good" driving, however, arguably two of the most important aspects are: (a) keeping the vehicle well positioned in the lane, and (b) driving the vehicle smoothly. 

\begin{figure}
    \centering
    \includegraphics[width=0.7\columnwidth]{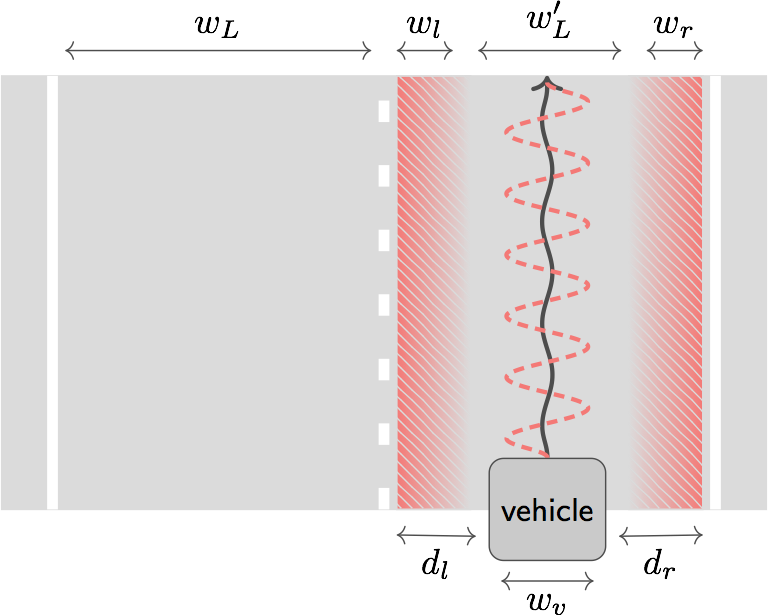}
    %\resizebox{0.7\columnwidth}{!}{\input{figures/penalty.tex}}
    \caption{In this figure, two lanes are indicated with the same lane width, $w_L$, where the penalty regions $w_l$ and $w_r$, non-penalized lane region $w_{L}^{'}$, vehicle width $w_v$ and vehicle--to--lane marking distances $d_l$ and $d_r$ are illustrated. The left and right lane markings are represented by white solid lines and the center lane marking is represented by a dashed white line.}
    \label{fig:performance_metric}
\end{figure}

A lane positioning error, $e_{\beta}(d)$ served the purpose of measuring the policy's performance in keeping the vehicle well aligned in the lane. Here, $d$ represents the distance between the vehicle and a lane marking. Figure~\ref{fig:performance_metric} illustrates a road segment with two lanes, both of width $w_L$, and a vehicle of width $w_v$ positioned in the right lane. The positioning penalty regions of the right lane are defined as the widths of the highlighted areas on either side of the vehicle, denoted $w_l$ and $w_r$ for the left and right regions respectively. The width of the non-penalized region of the lane is thus defined as $w_L' = w_L - (w_l + w_r)$. Further, the distances from the edges of the vehicle to the left and right lane markings are denoted $d_l$ and $d_r$ respectively.

The lane positioning penalty $e_{\beta}(d)$ is analogously defined for the left and right regions. Given a penalty region $w$ (either left or right) and vehicle to lane marking distance $d$ (corresponding left or right), the penalty function is defined~as

\begin{equation}
    e_{\beta}(d) = \begin{cases}
     1 & \text{ if $d < 0$}\\
     \left(\beta w\right)^{\frac{d}{w}} - \beta d & \text{ if $0 \leq d \leq w$} \\
     0 & \text{ if $d > w$}\\
    \end{cases}
\end{equation}
where $\beta$ is a hyper parameter that determines the shape of the error function w.r.t. the width of the penalty region $w$.

\begin{figure}
    \centering
    \includegraphics[width=0.7\columnwidth]{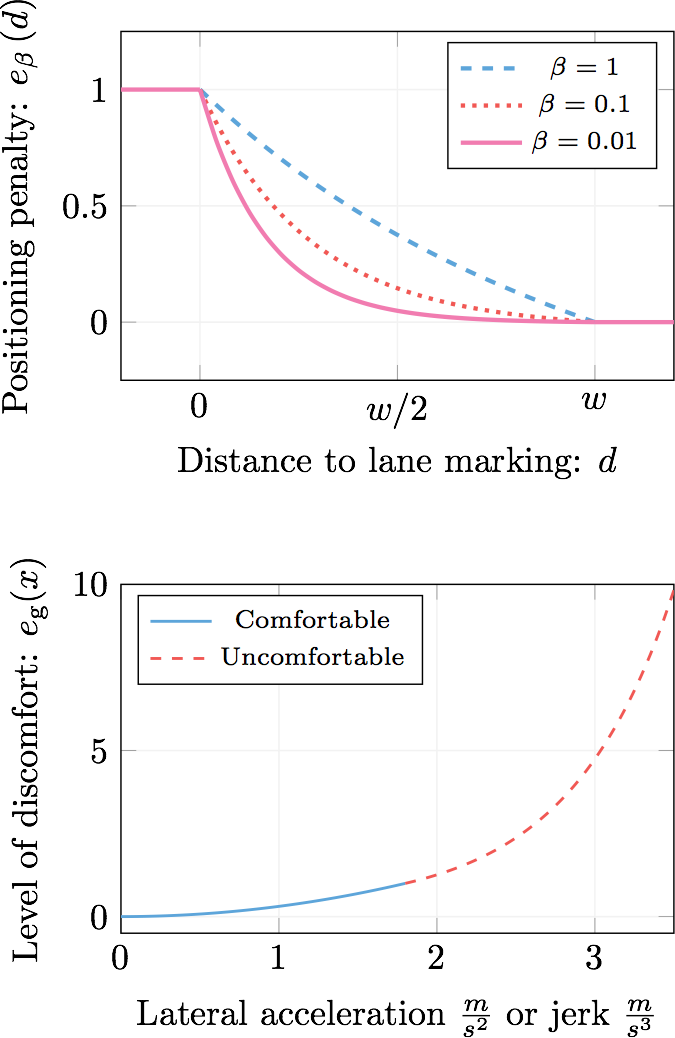}
    \caption{(Top plot) Lane positioning error as a function of vehicle--to--lane marking distance $d$ with respect to penalty region $w$ and $\beta$ for different choices of $\beta$. (Bottom plot) Level of discomfort metric $e_{g}(x)$ for a comfort threshold $g = 1.8$ m/s$^2$ or m/s$^3$ for $x$ being either lateral acceleration or jerk.}
    \label{fig:metrics}
\end{figure}

Figure~\ref{fig:metrics} illustrates the positioning penalty as a function of the vehicle to lane marking distance $d$ for three values of $\beta$. As can be seen, a value of $\beta$ close to 1 corresponds to an almost linear increase in error for decreasing distance to the lane marking while a smaller value of $\beta$ corresponds to an exponential increase in error as the vehicle gets closer to the lane marking. The choice of $\beta$ is subjective and depends on how one believes the penalty should grow as one approaches the lane markings. In either case, a value of 1 for the lane positioning penalty corresponds to entirely bad positioning while 0 corresponds to entirely good. 

In addition to vehicle positioning on the road, ``good" driving behavior in the sense of how smooth and comfortable a driven trajectory is, can be characterized by the level of lateral accelerations and jerks a vehicle is being subjected to. 
Two trajectories are illustrated in Figure~\ref{fig:performance_metric}, one as a solid line and the other as a dashed line in the right lane segment of the road. The solid line represents the better of the two trajectories where the vehicle is subject to less lateral accelerations and jerks than from the other. From empirical studies discussed in \cite{lennart-2011}, one can draw the conclusion that up to some certain threshold, jerks and accelerations only pose minuscule decreases in comfort. However, passing this threshold, the experienced discomfort rapidly increases. 

This level of discomfort (or error) $e_g(x)$ as a result of the acting acceleration and jerks is defined in \cite{lennart-2011} as
\begin{equation}
 e_g(x)= \begin{cases}
     \frac{x^2}{g^2} & \text{ if $x < g$}\\
     \left(\frac{5}{6}+\frac{x^2}{6g^2}\right)^6 & \text{ if $x\geq g$} \\
    \end{cases}
\end{equation}
where $x$ is either the measured lateral acceleration or jerk and $g$ is the comfort threshold. For this project, the comfort threshold was set to 1.8m/s$^2$ (or 1.8m/s$^3$) as it was deemed a good value based on empirical studies performed on highways in Canada and China \cite{felipe-1998,xu-2015}. Using this metric, a value of 0 would correspond to no accelerations at all (of course impossible in a turning vehicle) and a value of 1 would be on the very edge of what is deemed comfortable. Values above 1 corresponds to increasing levels of discomfort. An illustration of the level of discomfort can be seen in Figure~\ref{fig:metrics}.

\subsection{Test Scenarios}\label{sec:tests}
Scenarios used for empirical evaluation in the Unity game engine included driving on a six-lane highway with smooth curvature and a two-lane country road modelled as a much more winding road geometry from the Kajaman roads package~\cite{kajamansRoads}. For the highway, the speed limit was set to 100km/h while the country road was driven at speeds from 50-70km/h.

For quantitative evaluation, real road geometries obtained as latitude and longitude coordinates from Google Maps were used in the IPG CarMaker environment. The road geometries used corresponds to: (a) approximately 50km of Rikvs\"{a}g 160 from Rotviksbro to Ucklum in the south--west part of Sweden and (b) roughly 34km of the Volvo Drive Me route around Gothenburg, Sweden. The roads consist of 2 to 4 lanes, each of width 3.75m, and with speed limits varying from 50--90 km/h. The width of the simulated vehicle was approximately 2m.

In addition to evaluation of regular driving behaviour, the policy robustness and ability to recover from erroneous states was further empirically evaluated by the introduction of fault injections. That is, while driving, the steering wheel of the simulated vehicle was set so that the vehicle experienced misalignments with the road.

% =======================================================
% =====> EXPERIMENTAL RESULTS
% =======================================================

\section{Experimental Results}

\begin{figure}
    \centering
    \includegraphics[width=\columnwidth]{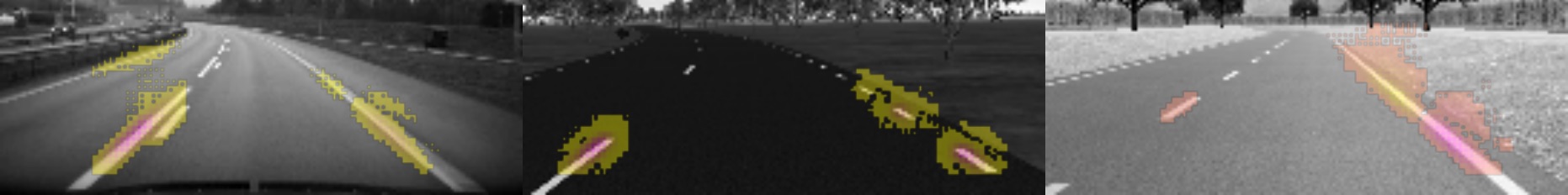}
    \caption{Illustration of the most significant pixel regions (pixels with the highest activation values), obtained through visual backpropagation from the policy trained on Volvo expedition data. Images are samples that correspond to a Volvo expedition (left), IPG CarMaker (center), and the Unity game engine (right).}
    \label{fig:visualbp}
\end{figure}

Empirical analysis on the effect of domain adaptation, as a result of the domain change imposed during evaluation, was initially carried out by identifying which pixel regions that contribute the most when making steering decisions. This was done by applying visual backpropagation (VBP)~\cite{visualbackprop2016} on state representations (images) from the real training data as well as synthetic data from the simulation environments and comparing these regions by visual inspection. Figure~\ref{fig:visualbp} illustrates a typical result of VBP and the similarities of important pixel regions across different domains. From this example, it is clear that the policy is able to react to similar concepts, i.e. lane markings, even though the model has never seen the synthetic data during training.

%  (illustrated in Figure~\ref{fig:carmaker_examples}),
% -------------------------------------------------------
% -----> TESTS AND EVALUATION
% -------------------------------------------------------

%Rating the performance of an AD system is not trivial. The performance needs to be quantified based on measures representing what good versus bad driving behaviour is, and as such, it's inherently subjective. 

%The metric proposed in \cite{dave2-2016}, denoted level of autonomy, is based on the amount of interventions needed during test drives. Applying this metric for this evaluation setup, results in a level of autonomy equal to 100\% (for both road geometries). This is not to say that the driving is perfect, but rather that the vehicle manages to stay in a lane throughout the entire test drive without interventions.

For this particular evaluation setup, using the real road geometries described in Section~\ref{sec:tests}, the lane penalty width can vary in the range $[0,0.875]$~m. Here 0~m corresponds to zero penalty being accumulated until the vehicle crosses either lane marking while 0.875~m corresponds to penalties being accumulated as soon as the vehicle deviates from the lane center. By measuring the distance to adjacent lane markings, it was noted that the vehicle maintained good positioning in the lane, staying roughly $0.5$~m from either lane marking 95\% of the time. By using the proposed lane penalty metric, and selecting a reasonable lane penalty width of approximately 0.4~m, the performance of the policy can be deemed good in terms of positioning approximately 99\% of the time. The erroneous 1\% stems from corner cutting in some of the sharper turns of the road segments and can be considered either good or bad depending on the situation, e.g. traffic, sight and road conditions.

In terms of smoothness, the penalty metric for level of discomfort was used to acquire a relative measure of smoothness. A relative measure is necessary because accelerations and jerks are highly dependent on road geometry and vehicle velocity. In the performed test, accelerations and jerks from an optimally driven trajectory was used for comparison. Optimal in this case means that the simulated vehicle traversed the same road geometries with identical velocity but the steering actions applied were obtained using the true curvature of the road geometries, i.e. the vehicle followed the absolute center of the lane at all times. 

Empirical results show that an optimal driver provides a roughly 1.1 times more comfortable trajectory than the learned policy in terms of acceleration, but almost 5 times as comfortable in terms of jerks.
The fact that the policy behaves considerably worse in terms of the level of jerk is not very surprising.
%This stems from the model property of making instantaneous decisions. The model simply does not take its previous decisions into account when making a new one, resulting in it rapidly shifting between different steering actions which in turn results in a jerkier trajectory\footnote{A simple solution that yielded significant improvements in terms of trajectory smoothness was by controlling the vehicle using an exponentially decaying average of the inferred steering actions as $\bar a_t \leftarrow \gamma a_t + (1-\gamma)\bar a_{t-1}$ with $\gamma \approx 0.1$.}.
This stems from the fact that the model makes instantaneous decisions, not taking previous decisions into account, which makes it rapidly shift between different steering actions\footnote{A simple solution that yielded significant improvements in terms of trajectory smoothness was by controlling the vehicle using an exponentially decaying average of the inferred steering actions as $\bar a_t \leftarrow \gamma a_t + (1-\gamma)\bar a_{t-1}$ with $\gamma \approx 0.1$. Here $a_t$ represents the steering action inferred by the driving policy at time $t$ and $\bar a_{t-1}$ is the filtered steering action of the previous time instance.}.
However, both accelerations and jerks obtained when using the policy as a driver lies, on average, within the comfortable range for the evaluated road geometries. The relative level of discomfort also seems to be stable and does not change significantly even if the road geometry is smooth or highly curved.

In addition, when disturbances are applied, such that the orientation of the vehicle suddenly becomes different from that of the road, the policy seems to have learned to correct the mistake.  To the best of our knowledge, no such corrective behaviour was included in the dataset, which then suggests that the policy was able to learn the concept of correcting from mistakes based solely on regular driving conditions. 

%This behaviour is of course highly dependent on the data used for learning and has been suggested not to work in \cite{dave2-2016} without data augmentation or otherwise including these scenarios in the dataset. 

The actual driving style of such a self-driving vehicle is still perhaps the best metric to evaluate the performance. Videos demonstrating our solution driving a vehicle in both Unity and CarMaker are available at
%\href{https://www.youtube.com/playlist?list=PLwaioBwah-xCZasmRHqEsvNfqTFgu9_Hf}
{goo.gl/MKKnuF}.

% =======================================================
% =====> CONCLUSION
% =======================================================

\section{Conclusion}
Based on the results obtained in this work, we conclude that: (a) a policy for lane keeping assistance learned by means of direct imitation learning using real data performs well in moderately different, never before seen simulated environments, (b) a sufficiently diverse dataset seems to provide a robust policy, and (c) using deep learning as a holistic solution to specific AD tasks such as lane keeping seems plausible even using simple sensor equipment such as a single front view camera. 
%In future work we will evaluate this solution in HIL and in real traffic environments. 

%\begin{figure}[t]
%\centering
%\includegraphics[width=\columnwidth]{cm-gray2.png}
%\framebox[0.7\columnwidth]{\large This is a figure!}
%\caption{Virtual front view images from  IPG CarMaker illustrating: (left) 2 lane road segment and (right) 4 lane road segment.}
%\label{fig:carmaker_examples}
%\end{figure}

%\newpage
% =======================================================
% =====> ACKNOWLEDGEMENT
% =======================================================

\section*{Acknowledgment}
The authors would like to thank Volvo Cars Corporation for sharing the test data and IPG Automotive for providing and supporting us with CarMaker.

% =======================================================
% =====> REFERENCES
% =======================================================



\bibliographystyle{IEEEtran}
\bibliography{IEEEabrv,database}
\end{document}